\documentclass[letterpaper]{article} 
\usepackage{aaai25}  
\usepackage{times}  
\usepackage{helvet}  
\usepackage{courier}  
\usepackage[hyphens]{url}  
\usepackage{graphicx} 
\def\UrlFont{\rm}  
\usepackage{natbib}  
\usepackage{caption} 
\usepackage{algorithm}
\usepackage{algorithmic}
\usepackage{color,xcolor}
\usepackage{colortbl}
\usepackage{multirow}
\usepackage{amsmath}
\usepackage{amssymb}
\usepackage{booktabs}

\usepackage{newfloat}
\usepackage{listings}
\DeclareCaptionStyle{ruled}{labelfont=normalfont,labelsep=colon,strut=off} 
\floatstyle{ruled}
\newfloat{listing}{tb}{lst}{}
\floatname{listing}{Listing}
\title{Query-centric Audio-Visual Cognition Network for \\ Moment Retrieval, Segmentation and Step-Captioning}
\author{
    Yunbin Tu\textsuperscript{\rm 1}, Liang Li\textsuperscript{\rm 2,\rm 1}\footnotemark[1], Li Su\textsuperscript{\rm 1,\rm 3}\footnotemark[1], Qingming Huang\textsuperscript{\rm 1}
}
\affiliations{
    \textsuperscript{\rm 1} School of Computer Science and Technology, University of Chinese Academy of Sciences, Beijing, China \\
    \textsuperscript{\rm 2} Key Laboratory of AI Safety of CAS, Institute of Computing Technology, Chinese Academy of Sciences, Beijing, China 
    \textsuperscript{3}Peng Cheng Laboratory, Shenzhen, China\\

    tuyunbin22@mails.ucas.ac.cn, liang.li@ict.ac.cn, \{suli,qmhuang\}@ucas.ac.cn
}

\usepackage{bibentry}

\begin{document}

\maketitle
\renewcommand{\thefootnote}{\fnsymbol{footnote}}
\footnotetext[1]{Corresponding authors.}

\begin{abstract}
Video has emerged as a favored multimedia format on the internet. To better gain video contents, a new topic HIREST  is presented, including video retrieval, moment retrieval, moment segmentation, and step-captioning. The pioneering work chooses the pre-trained CLIP-based model for video retrieval, and leverages it as a feature extractor for other three challenging tasks solved in a multi-task learning paradigm. Nevertheless, this work struggles to learn the comprehensive cognition of user-preferred  content, due to disregarding the hierarchies and association relations across modalities. In this paper, guided by the shallow-to-deep principle, we propose a  query-centric audio-visual cognition (QUAG) network to construct a reliable multi-modal representation for moment retrieval, segmentation and step-captioning. Specifically, we first design the modality-synergistic perception to obtain rich audio-visual content, by modeling global contrastive alignment and local fine-grained interaction between visual and audio modalities. Then, we devise the query-centric  cognition that uses the deep-level query to perform the temporal-channel filtration on the shallow-level audio-visual representation.  This can cognize user-preferred content and thus attain a query-centric audio-visual representation for three  tasks. Extensive experiments show   QUAG  achieves the SOTA results on HIREST. Further, we test QUAG on the query-based video summarization task and verify its good generalization. The code is available at \url{https://github.com/tuyunbin/QUAG}.
\end{abstract}

%

\section{Introduction}

\begin{figure}[t]
  \centering
   \includegraphics[width=1\linewidth]{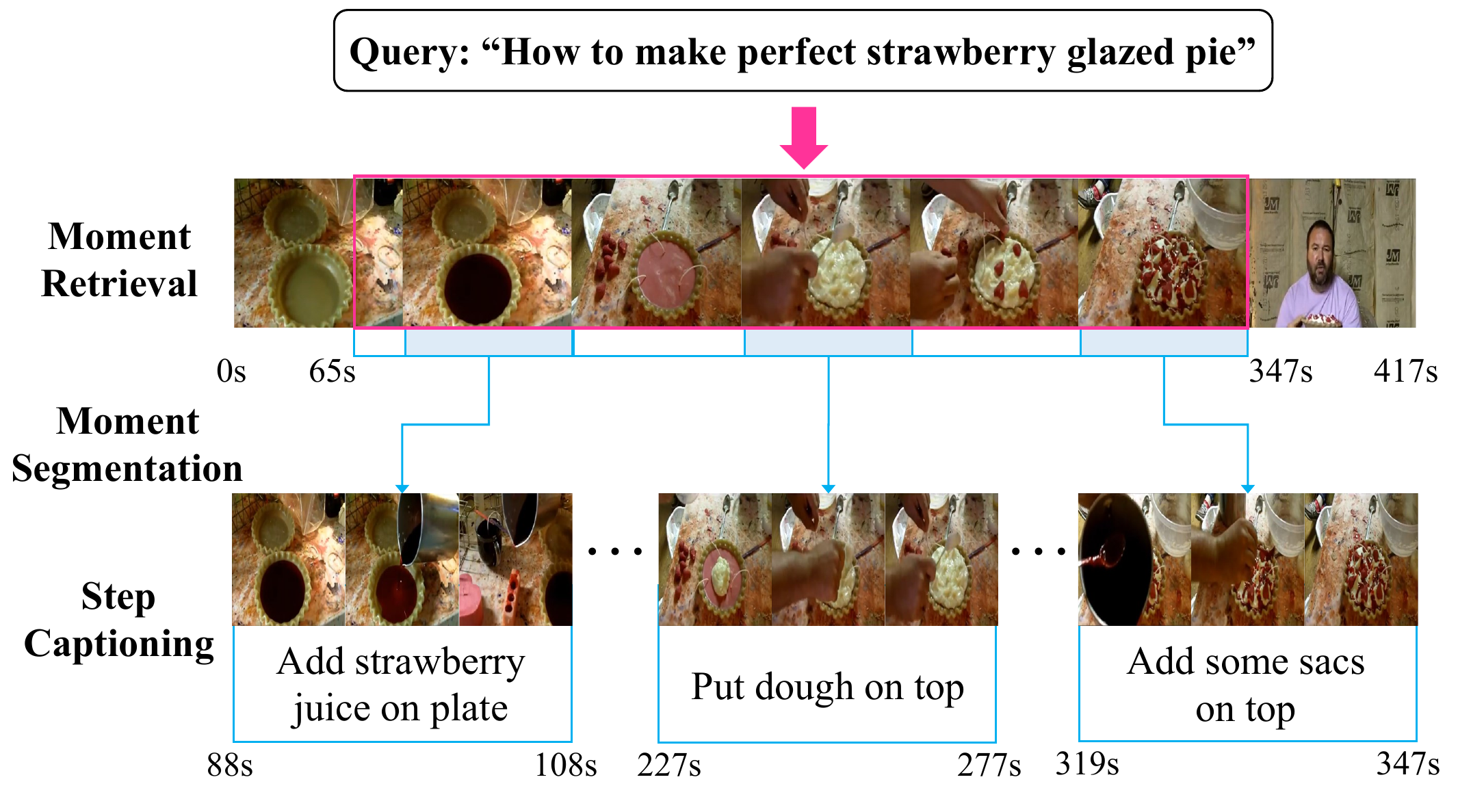}
   \caption{The illustrative example consisting of moment retrieval, moment segmentation, and step-captioning. First, given a text query ``How to make perfect strawberry glazed pie'', the model is required to localize the most related moment in the video (moment retrieval). Then, the model proceeds to break down the moment into finer-level steps (moment segmentation). Finally, the model should  describe each step with a concise sentence (step-captioning).
   }
   \label{fig1}
\end{figure}

Recently, we have been witnessing an exponential growth of videos, along with the advance of generative  AI (\emph{e.g.,} Sora) and video platforms. 
Accordingly, a plethora of studies \cite{wu2023cap4video,li2024text,wang2024dual} have been proposed to enhance video retrieval capabilities. However, given a text query, returning the whole video is not always satisfied. Sometimes, users  want to directly localize the moment most related to the query. For instance,  when learning how to make a strawberry pie, users tend to only focus on the instructional moment, as shown in Figure \ref{fig1}.
Moreover, if this moment is intricate, we anticipate machines to segment it into finer-level steps, and caption each with a succinct sentence for a better understanding, such as ``Add strawberry juice on plate'', ``Put dough on top'', etc. This shift in user preference sparks an emerging research
 topic: hierarchical retrieval and step-captioning (HIREST)  \cite{zala2023hierarchical}.

This topic benchmarks four tasks of video retrieval, moment retrieval,  moment segmentation, and  step captioning. In the pioneering work \cite{zala2023hierarchical}, Zala \emph{et al.} implement video retrieval by the pre-trained CLIP-based model and use it as a feature extractor for the three downstream tasks. Their focus is to jointly address the three tasks in a single architecture. Toward this end, they propose a \textit{Joint} model that first leverages pre-trained models \cite{fang2023eva,radford2023robust, reimers2019sentence} to produce the multi-modal representation for  visual frames, audio and query. Next, this representation is used to predict  the boundaries of moments and steps, as well as yielding each step caption. In this process, the  model is trained via a multi-task setup in a round-robin way \cite{cho2021unifying}.

Despite the promising results,  the above work has the limitations in learning a comprehensive cognition of user-preferred video content, owing to the naive multi-modal fusion strategy. 
First, the representations of three modalities are directly fused without distinction, which disregards the hierarchies across modalities.  Studies in psychology \cite{tacca2011commonalities,yang2021multiple} have revealed that human perception and cognition is in a shallow-to-deep fashion. When  watching a video, we usually first obtain shallow-level sensory information, \emph{e.g.,} the appearance and sound of  objects, to obtain an intuitive perception for audio-visual content in the video. Then, we incorporate deeper-level knowledge (\emph{e.g.,} intentions) into the sensory information to learn the comprehensive cognition of the particular content of interest. Hence, it is beneficial to model the modality hierarchies based on this shallow-to-deep principle. 

Second, the fusion strategies (element-wise multiplication and summation) among three modalities fail to make full use of their association relations. %
The shallow-level visual and audio modalities can represent video content from different aspects, but direct element-wise summation may lose their synergistic relation. Afterwards, the deep-level  query can filter trivial details and highlight the important ones within the shallow-level audio-visual content. Nevertheless, element-wise multiplication cannot model such a filtration relation  that helps cognize user-preferred  video content. As such, it is warranted to progressively capture these association relations during modeling hierarchies across modalities.

In this paper, we propose a  \textbf{QU}ery-centric \textbf{A}udio-visual co\textbf{G}nition (QUAG) network to cognize user-preferred  video content and learn an effective  multi-modal representation for addressing moment retrieval, segmentation and step-captioning. QUAG consists of  modality-synergistic perception (MSP) and query-centric  cognition (QC$^2$) modules. Concretely, given visual frame and audio representations,  MSP first models their global contrastive alignment to make them reside in the same embedding space; then learns their local fine-grained interaction to mine their joint representations, which are fused as the audio-visual representation. Afterwards, guided by the deep-level query,  QC$^2$ performs a temporal-channel filtration on the shallow-level audio-visual representation, thus highlighting user-requested details. Next, the query representation is injected into the filtered audio-visual representation to construct a query-centric audio-visual representation. This is finally fed into the multi-modal encoder, multi-task prediction heads and text decoder for addressing the three challenging tasks.

Our key contributions are summarized as follows:
\begin{itemize}
    \item Based on the investigation for human perception and cognition, we follow the shallow-to-deep principle to propose QUAG, which jointly solves  three challenging tasks via learning a query-centric audio-visual representation.

    \item In QUAG, we first design MSP to attain the audio-visual representation by modeling global contrastive alignment and local fine-grained interaction between visual and audio representations. Then, we devise QC$^2$ to attend to the query-centric audio-visual representation, by implementing temporal-channel filtration from the deep-level query  to  the shallow-level  audio-visual representation.

    \item Extensive experiments show  QUAG  achieves state-of-the-art results for the moment retrieval, segmentation and step-captioning   on the HIREST dataset. Moreover, we test QUAG on the TVSum dataset for the query-based video summarization, which validates its generalization. 
\end{itemize}

\section{Related Work}
The task of hierarchical retrieval and step-captioning belongs to the multi-modal learning community \cite{cong2022ls,cong2023learning,tu2023self,tu2024distractors,tu2024smart,tu2024context}. In the following, we review the related works from four dimensions. 

\textbf{Video-moment retrieval.} Cross-modal retrieval is a fundamental problem in artificial intelligence \cite{zha2019context,zhang2020causal,liu2022entity,zhang2024inductive,yue2023i3n,wang2024zero,wang2024dets,tang2024context,yue2024multi}.  Text-based video retrieval \cite{dong2022partially,xie2024phrase,li2023progressive} is to find related videos from a  corpora by a text query. Recently, benefiting from the contrastive learning, most text-to-image/video retrieval models (\emph{e.g.,} CLIP) have been designed \cite{radford2021learning,fang2023eva}. With these models, the cosine similarity between the text query and video can be easily computed, thus finding the most related video to the query. However, since there are some query-irrelevant parts in the whole video,  moment retrieval has been studied by most methods \cite{sun2022you,lei2021detecting,moon2023query}, which is to localize the query-related span  in the video. 


\textbf{Video summarization.} Traditional video summarization methods \cite{song2015tvsum,xiong2014detecting} are to condense lengthy videos by extracting important information.  However, these methods ignore the users' various preferences  over the summaries. To address this limitation, the query-focused approaches \cite{sharghi2016query,sharghi2017query,narasimhan2022tl} incorporate users' preferences through text queries, thus identifying the most relevant frames within the video. By aligning the summarization process with user interests, these query-focused methods can find a collection of moments that are most related to the users' preferences.


\textbf{Video Captioning.} Traditional video captioning methods \cite{li2022long,tu2023relation} generate a concise sentence for a short video. To describe longer videos with multiple events, dense video captioning is studied by some methods \cite{yang2023vid2seq,kim2024you}, which is to generate a paragraph for the  long video. On the other hand, there are some methods \cite{lei2020tvr,li2020hero,tu20222} that try to study TV show captioning. Instead of only using visual modalities,  they  consider introducing text features (\textit{e.g.}, subtitles from the actors' dialogues) to augment video features, thus modeling a visual-linguistic representation for caption generation.

\begin{figure*}[t]
\centering
\includegraphics[width=1\textwidth]{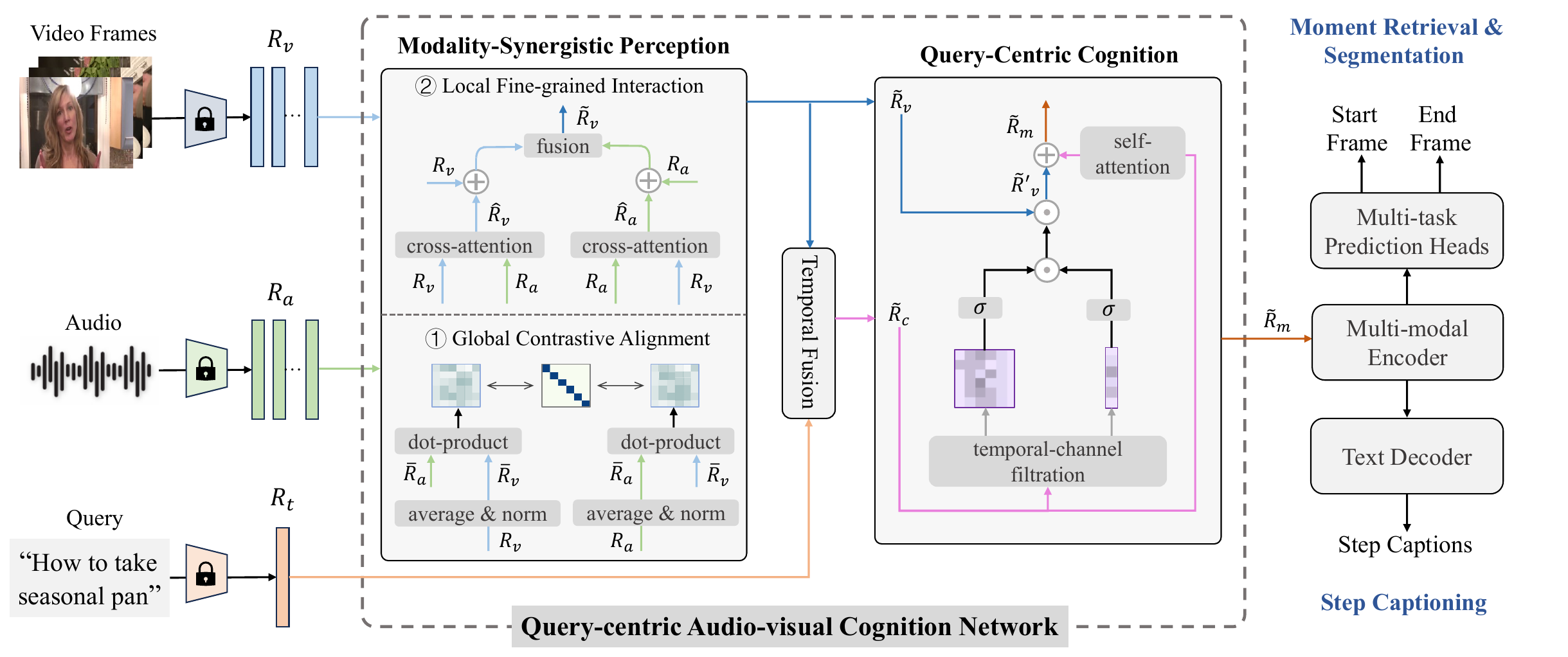} 
  \caption{The overview of our method. Based on the principle of shallow-to-deep, we propose a query-centric audio-visual cognition (QUAG) network, where  the core modules are the \textbf{modality-synergistic perception} and \textbf{query-centric cognition}. QUAG aims to learn a comprehensive cognition of user-preferred video content, and thus attain a query-centric  audio-visual representation for jointly addressing the moment retrieval, moment segmentation, and step-captioning. }
\label{fig2}
\end{figure*}

\textbf{Hierarchical Retrieval and Step-Captioning.} The above  tasks are individually studied by previous works. In fact, these tasks share a common goal to extract information from a video corpus. 
Thus, Zala \emph{et al.} \cite{zala2023hierarchical} combine these tasks as a new task called HIREST to cater to users' various preferences. In this pioneering work, 
the pre-trained CLIP-based model is leveraged for video retrieval, and used as the feature extractor for the other three tasks. Further, a unified framework is proposed to jointly address the three tasks. In this work, we follow this paradigm to address the three tasks in a unified architecture. Compared to the pioneering work that directly fuses multiple modalities by element-wise summation/multiplication, we learn a query-centric audio-visual representation by modeling the hierarchies and association relations across  modalities, so as to better solve the three downstream tasks. 

\section{Methodology}
As shown in Figure \ref{fig2}, the overall framework of our method includes the following parts. First, given the visual frames,  audio,  and  query, we first extract the visual, audio, and text representations by  pre-trained models. Then, we feed them into  QUAG to produce a query-centric audio-visual representation. Finally, this representation is fed into a multi-modal encoder for intra-relation modeling, and used to 1) predict the boundaries of moments and steps, and 2) prompt the text decoder to generate the caption for each step.  

\subsection{Multi-modal Input Embedding}
Given an untrimmed video  containing $N_v$ frames and a text query  consisting of $N_t$ tokens, we first employ pre-trained models to extract the visual representation $R_v$ and text representation $R_t$, separately. Then, considering that the audio information helps perceptive the main objects in the video, we also extract the speech transcription from the audio and embed it as the audio representation $R_a$, whose length is equal to the video representation, \emph{i.e.,} $N_v$. Next, we project three modality representations into the same embedding space by three linear transformation functions, \emph{i.e.,} $R_v \in \mathbb{R}^{N_v \times D}$, $R_a \in \mathbb{R}^{N_v \times D}$, $R_t \in \mathbb{R}^{D}$.

\subsection{Query-centric Audio-visual Cognition}
After obtaining the visual, audio, and text representations, we propose a query-centric audio-visual cognition (QUAG) network that utilizes the modality-synergistic perception (MSP) and query-centric cognition (QC$^2$) modules to learn a query-centric audio-visual representation based on the shallow-to-deep principle, so as to learn the comprehensive cognition of user-preferred  video content. 

\subsubsection{Modality-Synergistic Perception} 
Given the pair-wise visual representation $R_v$ and audio representation $R_a$, we design the MSP to first make them reside in the same embedding space by maximizing their global contrastive alignment. Specifically, we compute their global features by the mean-pooling operation over the length dimension:
\begin{equation}
\bar R_v =\frac{1}{N_v} \sum_{N_v} {R_v}, \bar R_a =\frac{1}{N_v} \sum_{N_v} {R_a},
\end{equation}
Where $\bar R_v \in \mathbb{R}^{D}$ and $\bar R_a \in \mathbb{R}^{D}$. Then, in a training batch, we sample $B$ pair-wise globally visual and audio features. For $k$-th globally visual feature  $\bar{R}_v^k$,  $k$-th globally audio feature $\bar{R}_a^k$ is its positive, while other $r (r \neq k)$ globally audio features will be the negatives. That is, there are a total of 
 $B$ positive sample pairs and $B \times (B-1)$ negative sample pairs in this batch. 
Next, we employ the InfoNCE loss \cite{oord2018representation} to maximize the bi-directional similarities between the positive pairs $\bar{R}_v^k$ and $\bar{R}_a^k$, while  minimizing the similarity between the negative pairs:
\begin{equation}
\label{infonce}
\begin{gathered}
\mathcal{L}_{v2a}=-\frac{1}{B} \sum_k^B \log \frac{e^{ \left(\text{sim}\left( \bar{R}_v^k, \bar{R}_a^k\right) / \tau\right)}}{\sum_r^B e^{ \left(\text{sim}\left(\bar{R}_v^k,  \bar{R}_a^r\right) / \tau\right)}}, \\
\mathcal{L}_{a2v}=-\frac{1}{B} \sum_k^B \log \frac{e^{ \left(\text{sim}\left( \bar{R}_a^k, \bar{R}_v^k\right) / \tau\right)}}{\sum_r^B e^{ \left(\text{sim}\left(\bar{R}_a^k,  \bar{R}_v^r\right) / \tau\right)}},\\
\mathcal{L}_{msp}=\frac{1}{2}(\mathcal{L}_{v2a}+\mathcal{L}_{a2v}),
\end{gathered}
\end{equation}
where $\tau$  is the temperature hyper-parameter. ``sim'' means the dot-product operation.  This loss formulates a self-supervisory signal to enhance the global alignment between the visual and audio representations, which facilitates the subsequent local fine-grained interaction. 

Subsequently, we interact the local features on the visual representation $R_v$ and audio representation $R_a$ to learn their fine-grained synergy and mine the joint representations. This  is performed by the multi-head cross-attention mechanism (MHCA) \cite{vaswani2017attention}:
\begin{equation}
\begin{gathered}
\hat R_v=\operatorname{MHCA}\left(R_v, R_a, R_a\right), \\
\hat R_a=\operatorname{MHCA}\left(R_a, R_v, R_v\right).
\end{gathered}
\end{equation}
In the bracket of each equation, the first term represents the query, while the last two terms denote the key and value representations.
After that, we concatenate the joint representations from two modalities as the audio-visual representation, which is performed by a fully-connected layer:
\begin{equation}
\tilde{R}_{v}=\left(\left[\hat R_v ; \hat R_a\right] W_{c} + b_c \right ),
\end{equation}
where $\tilde R_v \in \mathbb{R}^{N_v \times D}$. $W_{c}\in \mathbb{R}^{2D \times D}$ and $b_{c}\in \mathbb{R}^{D}$.  [;] is a concatenation operation. 

\subsubsection{Query-Centric Cognition} We devise QC$^2$ to use the deep-level  query representation  $R_t \in \mathbb{R}^{D}$ to attend to the related information over the shallow-level audio-visual representation $\tilde R_v \in \mathbb{R}^{N_v \times D}$. QC$^2$ first measures the relevance between $R_t$ and $\tilde R_v$ from both temporal and channel dimensions. The temporal and channel relation matrices $\mathbf{A}_{\mathbf{t e}}$ and $\mathbf{A}_{\mathbf{ch}}$ are computed as follows:
\begin{equation}
\begin{aligned}
& \quad \quad \quad \tilde{R}_{c}=\left(\left[\tilde R_v ;  R_t\right] W_{d} + b_d \right ), \\
& \mathbf{A}_{\mathbf{t e}}=\operatorname{sigmoid}\left(\mathcal{F}_{t e}\left(\frac{1}{D} \sum_j^{D} \tilde R_c (:, j)\right)\right), \\
& \mathbf{A}_{\mathbf{c h}}=\operatorname{sigmoid}\left(\mathcal{F}_{c h}\left(\frac{1}{N_v} \sum_j^{N_v}  \tilde R_c(j,:)\right)\right), 
\end{aligned}
\end{equation}
where $W_{d}\in \mathbb{R}^{2D \times D}$ and $b_{d}\in \mathbb{R}^{D}$.  [;] is a concatenation operation and we broadcast the text representation $R_t \in \mathbb{R}^{D}$ as $R_t \in \mathbb{R}^{N_v \times D}$,  to temporally match its shape with the audio-visual representation $ \tilde R_v \in \mathbb{R}^{N_v \times D}$.  $\mathbf{A}_{\mathbf{t e}} \in \mathbb{R}^{N_v \times 1}$, $\mathbf{A}_{\mathbf{ch}} \in \mathbb{R}^{1 \times D}$. $\mathcal{F}_{t e} \in \mathbb{R}^{1 \times 1}$ and $\mathcal{F}_{ch} \in \mathbb{R}^{D \times D}$ are linear transformation functions. 

Next, QC$^2$ computes the temporal-channel relation matrix by fusing $\mathbf{A}_{\mathbf{t e}}$ and $\mathbf{A}_{\mathbf{ch}}$ together, which is implemented with the element-wise multiplication function:
\begin{equation}
\mathbf{A}_{\mathbf{tc}}=\mathbf{A}_{\mathbf{te}} \odot \mathbf{A}_{\mathbf{c h}},
\end{equation}
where $\mathbf{A}_{\mathbf{t c}} \in \mathbb{R}^{N_v \times D}$. Guided by $\mathbf{A}_{\mathbf{t c}}$, the model can filter the trivial details in the audio-visual representation $\tilde{R}_{v}$: 
\begin{equation}
\tilde R'_{v} =\mathbf{A}_{\mathbf{tc}} \odot \tilde{R}_{v},
\end{equation}
where $\tilde R'_{v} \in \mathbb{R}^{N_v \times D}$. Subsequently, we  integrate $\tilde{R}_{c}$ into the filtered audio-visual representation $\tilde R'_{v}$ to model a query-centric audio-visual representation:
\begin{equation}
 \tilde R_{m} = \tilde{R'}_{v} + \phi (\tilde{R}_{c}),
\end{equation}
where $\phi$ refers to the  multi-head self-attention.
Finally, $\tilde R_{m}$ is fed into a transformer-based multi-modal encoder for enhancement, and then is utilized for the moment retrieval, moment segmentation and step-captioning.

\subsection{Moment Retrieval, Moment Segmentation, and Step-captioning}
\textbf{Moment Retrieval} is to find the moment most relevant to the query and output its span. Specifically, given $\tilde R_{m}$, we use a learnable prediction head consisting of two linear layers to predict the start and end boundaries at the same time, where the  frame inputs are not masked. The probability distributions of the start and end indexes over the entire  video are computed as follows:
\begin{equation}
\label{word}
\begin{gathered}
    \mathbf{P_{start}}=\operatorname{Softmax}\left(\tilde R_{m} W_{s}+{b}_{s}\right), \\
    \mathbf{P_{end}}=\operatorname{Softmax}\left(\tilde R_{m} W_{e}+{b}_{e}\right),
\end{gathered}
\end{equation}
where $W_{s}\in \mathbb{R}^{D \times 1 }$, $W_{e}\in \mathbb{R}^{D \times 1 }$, $b_{s} \in \mathbb{R}^{1 }$, and $b_{e} \in \mathbb{R}^{1 }$. 
 
\noindent \textbf{Moment Segmentation} refers to identifying ``key steps'' within the retrieved moment that does not have repetitive frames, where each step’s end timestamp is the next step’s start timestamp. Concretely, given $\tilde R_{m}$, we use a learnable prediction head with a linear layer to predict the boundary of each step in an auto-regressive manner, where the previously predicted boundaries are integrated into $\tilde R_{m}$  as the part of the input. The frames outside of the moment and in the previous steps are masked out. The probability distribution of boundary index for each step is computed as follows:
\begin{equation}
\label{word}
\begin{gathered}
    \mathbf{P_{step}}=\operatorname{Softmax}\left(\tilde R_{m} W_{t}+{b}_{t}\right), \\
\end{gathered}
\end{equation}
where $W_{t}\in \mathbb{R}^{D \times 1 }$ and $b_{e} \in \mathbb{R}^{1 }$.

\noindent \textbf{Step-captioning} aims to describe each segmented step with a sentence. To this end,  we first use the pre-trained captioning model CLIP4Caption \cite{tang2021clip4caption} to initialize the parameters of text decoder. Then, we feed  $\tilde R_{m}$ into the text decoder that yields the text descriptions through an auto-regressive way. This process can be formulated as follows:
\begin{equation}
\label{word}
\hat {\mathbf{w}_{i}}=\mathcal{F}_{D e}\left(\tilde R_{m}, {\mathbf{w}}_1, \ldots, {\mathbf{w}}_{i-1}\right),
\end{equation}
where $\hat{\mathbf{w}}_i \in \mathbb{R}^{N_t}$ is the $i$-th decoded distribution over a dictionary of $N_t$ vocabularies; $\left\{{\mathbf{w}}_1, \ldots, {\mathbf{w}}_{i-1}\right\}$ are previously generated  words (ground-truth words during training and predicted words during inference). 

\noindent \textbf{Training.} For moment retrieval, given the ground-truth start and end indices, we define the training loss (to be minimized) as the sum of the negative log probabilities by the predicted distributions, which are then averaged over all examples in a batch:
\begin{equation}
\mathcal{L}_{ret}=-\frac{1}{B} \sum_i^B \log \left(\mathbf{p}_{start}^{y_i^{start}}\right)-\frac{1}{B} \sum_i^B\log \left(\mathbf{p}_{end}^{y_i^{end}}\right),
\end{equation}
where $B$ is the batch size; $y_i^{start}$ and $y_i^{end}$ are the ground-truth start and end indices of the $i$-th example, respectively.

For moment segmentation, given ground-truth step indices, we define the training loss (to be minimized) as the sum of the negative log probabilities by the predicted distributions, which are averaged over all examples in a batch:
\begin{equation}
\mathcal{L}_{seg}=-\frac{1}{B} \sum_i^B \log \left(\mathbf{p}_{step}^{y_i^{step}}\right),
\end{equation}
where $B$ is the batch size  and $y_i^{step}$ is the ground-truth step index of the $i$-th example, respectively.

For step-captioning, given the ground-truth step caption words $\left(w_{1}^{*}, \ldots, w_{m}^{*}\right)$, we define the training loss (to be minimized) as the sum of the negative log probabilities by the predicted distributions, which are then averaged over all examples in a batch:
\begin{equation}
\mathcal{L}_{{cap }}=-\sum_{i=1}^m \log p_\theta\left(w_i^* \mid w_{<i}^*\right)
\end{equation}
where $p_\theta\left(w_i^{*} \mid w_{<i}^{*}\right)$ is computed by Eq.~(\ref{word}). $m$ is the length of a step caption.

Our method is trained in a multi-task configuration, where a round-robin manner is leveraged and  a batch is sampled from one of the three data loaders at each iteration \cite{zala2023hierarchical,cho2021unifying}. That is, the training loss for each iteration is defined as:
\begin{equation}
\label{loss}
\mathcal{L}(\theta)= \mathcal{L}_i + \lambda  \mathcal{L}_{msp},
\end{equation}
where $i$ refers to $ret$, $seg$, or $cap$. $\theta$ are a set of trainable weights. $\lambda$ is a trade-off parameter to balance the contribution between one of the task losses and the contrastive loss, which is discussed in the supplementary material \footnote{The supplementary material is included in the arXiv version.}.

\section{Experiments}
\subsection{Datasets}

\noindent \textbf{HIREST} consists of the tasks of video retrieval, moment retrieval, moment segmentation, and step-captioning. It is comprised of 3.4K text-video pairs, 1.8K moments, and 8.6K step captions. 
We use the official split with 1,507 video-query pairs for training, 477 video-query pairs for validation and 1,391 video-query pairs for testing.


\noindent \textbf{TVSum} contains  the task of query-based
 video summarization, which is relevant to  moment segmentation. This dataset includes 10 varying categories of videos, and each category consists of 5 videos. For a fair-comparison, we follow QD-DETR \cite{moon2023query} to utilize 80\% videos for training and the remaining for testing. 

\subsection{Evaluation Metrics}
\textbf{HIREST:} (1) We validate moment retrieval by  evaluating the outputs of model against the ground-truth moment spans via Recall of Intersection over Union (IoU) thresholds (0.5 and 0.7).
(2) For moment segmentation, models are assessed based on the similarity between the generated step spans and the ground-truth  step spans, via Recall and Precision metrics over IoU thresholds (0.5 and 0.7).
(3) For step-captioning, the traditional metrics are used to evaluate the yielded sentence:  METEOR \cite{banerjee2005meteor}, ROUGE-L \cite{lin2004rouge}, CIDEr \cite{vedantam2015cider}, and SPICE  \cite{anderson2016spice}. Besides, Following \cite{zala2023hierarchical}, we introduce the ELMo \cite{peters2018deep}-based Decomposable Attention model \cite{parikh2016decomposable} to compute the entailment of generated sentences against the ground-truth captions.



\noindent \textbf{TVSum:} Following previous works \cite{moon2023query,liu2022umt},
we leverage the top-5 mAP as the main metric.

\subsection{Implementation Details}
 First, we use pre-trained EVA-CLIP \cite{fang2023eva} to extract the visual representation for  visual frames, and the text encoder of EVA-CLIP to map a text query as the text representation. Then, we use pre-trained Whisper \cite{radford2023robust} to extract speech transcription from audio, and use pre-trained MiniLM \cite{reimers2019sentence} text encoder to map the transcription into a representation, called audio representation. The hidden size is set to 768. The text decoder is first initialized from pre-trained CLIP4Caption \cite{tang2021clip4caption}, and then is fine-tuned on the HIREST dataset. During training, the batch size is set to 5 and learning rate is set to 1 $\times$ $10^{-5}$, AdamW optimizer \cite{loshchilov2018decoupled} is used to minimize the training loss defined in Eq. (\ref{loss}). More details are shown in supplementary. 

\subsection{Performance Comparison on HIREST}

\subsubsection{Results on the Moment Retrieval} In this task, we compare QUAG with the  pioneering work \textit{Joint} model \cite{zala2023hierarchical}. We also compare QUAG with three task-specific moment retrieval models: QD-DETR   \cite{moon2023query}, TR-DETR \cite{sun2024tr}, and UVCOM \cite{xiao2024bridging}. The mentioned  models are implemented based on their released codes. Besides, following Zala \emph{et al.} \cite{zala2023hierarchical}, we compare QUAG with a  text-to-image retrieval model EVA-CLIP \cite{fang2023eva}. 

The comparison results are shown in Table \ref{moment retrieval}. Our QUAG obtains the best  performances in moment retrieval at both metrics. These indicate our method has the high recall in identifying relevant moments with the 50\% overlap and 70\% overlap with the ground-truth spans. Especially, our method has the performance improvements of 2.2\% and 4.2\% against the  \textit{Joint} model at R@0.5 and R@0.7, respectively, which further validates the effectiveness of  QUAG. 

\begin{table}[htbp]
  \centering
  \caption{ Moment retrieval results on the HIREST test set.}
    \begin{tabular}{c|cc}
    \toprule
    \multirow{2}[4]{*}{Model} & \multicolumn{2}{c}{Recall @ IoU} \\
\cmidrule{2-3}          & 0.5   & 0.7 \\
    \midrule
    EVA-CLIP-G/14  \cite{fang2023eva} & 38.27 & 19.33 \\ \midrule
    QD-DETR  \cite{moon2023query} & 71.52 & 38.34 \\
    UVCOM \cite{xiao2024bridging}  & \underline{72.12} & \underline{38.53} \\
    TR-DETR \cite{sun2024tr} & {72.07} & 38.40 \\ \midrule
    \textit{Joint} \cite{zala2023hierarchical} & 70.98 & 37.31 \\
    \textbf{QUAG } & \textbf{72.54} & \textbf{38.86} \\
    \bottomrule
    \end{tabular}%
  \label{moment retrieval}%
\end{table}%

\begin{table}[t]
  \centering
  \caption{Moment segmentation results on the test set of the HIREST dataset.}
    \begin{tabular}{c|cc|cc}
    \toprule
    \multirow{2}[4]{*}{Model} & \multicolumn{2}{c|}{Recall @ IoU} & \multicolumn{2}{c}{Precision @ IoU} \\
\cmidrule{2-5}          & 0.5   & 0.7   & 0.5   & 0.7 \\
    \midrule
    QD-DETR  & 34.24 & 16.27  & 28.32 & 12.05 \\
    UVCOM   & 37.03 & 16.79  & \underline{30.38} & \underline{13.80} \\
    TR-DETR  & 37.07 & 16.35 & 29.71 & 12.93 \\ \midrule
    \textit{Joint}  & \underline{36.23} & \underline{16.05} & 28.52 & 12.84 \\
    \textbf{QUAG} & \textbf{39.27} & \textbf{17.35} & \textbf{31.68} & \textbf{14.81} \\
    \bottomrule
    \end{tabular}%
  \label{moment segmentation}%
\end{table}%

\begin{table}[htbp]
\small
  \centering
  \caption{Step-captioning results on the HIREST test set. M, R, C, S indicate METEOR, ROUGE-L, CIDEr, and SPICE. }
    \begin{tabular}{c|ccccc}
    \toprule
    Model & M & R & C & S & Entail. (\%) \\
    \midrule
    BMT & 3.84  & -     & 6.72  & 1.05  & 30.68 \\
    SWinBERT & \textbf{5.94} &   -    & 24.66 & \textbf{6.67} & 35.09 \\ \midrule
    \textit{Joint} & 4.14  & 11.85 & 21.19 & 3.02  & 35.97 \\
    \textbf{QUAG} & \underline{4.76}  & \textbf{13.20} & \textbf{25.44} & \underline{4.49}  & \textbf{40.10} \\
    \bottomrule
    \end{tabular}%
  \label{step-captioning}%
\end{table}%

\subsubsection{Results on the Moment Segmentation} In this task, we compare the proposed QUAG with the  \textit{Joint} model \cite{zala2023hierarchical}. Also, three task-specific models QD-DETR \cite{moon2023query}, TR-DETR \cite{sun2024tr}, and UVCOM \cite{xiao2024bridging} are implemented based on their codes and compared with our QUAG. 

The results are shown in Table \ref{moment segmentation}. The QUAG excels in Recall at IoU thresholds of 0.5 and 0.7. QUAG’s high recall rates at both IoU thresholds shows its strong ability for moment segmentation. For the precision metric, our QUAG obtains the best result at IoU thresholds of both  0.5 and 0.7, which indicates our method can segment a moment with a high degree of accuracy. Overall, the superior  results on both recall and precision demonstrate QUAG’s effectiveness in achieving  accurate and reliable moment segmentation, which is essential for the subsequent step-captioning task.

\subsubsection{Results on the Step-captioning} In this task, we compare the proposed QUAG with the \textit{Joint} model \cite{zala2023hierarchical}, BMT \cite{iashin2020better} that is a dense video captioning model  pre-trained on the ActivityNet dataset \cite{krishna2017dense}), and SwinBERT \cite{lin2022swinbert} that is a pre-trained video captioning model on the YouCook2 dataset \cite{zhou2018towards}. 
The comparison results are shown in Table \ref{step-captioning}. 

It is noted that the overall performance of our QUAG is better in this task. Our QUAG outperforms the \textit{Joint} model on every metric by a large margin.  For two captioning-based models BMT and SwinBERT, our QUAG  outperforms BMT on all metrics and achieves the matching performance against SwinBERT. Especially, on the caption-centric metric CIDEr, our QUAG obtains the relative improvements of 3.2\%
and 280\% over SwinBERT and BMT, respectively. On the entailment score, QUAG also obtains the relative improvements of 14.3\% and 30.7\% over SwinBERT and BMT, separately. The comparison results show the good generalization of our method for caption generation. 

\textbf{Overall Performance Analysis.} Compared with the pioneering work \textit{Joint} on HIREST,  our method surpasses it in every metric for each sub-task, proving its effectiveness. To comprehensively evaluate our approach, it is also compared with some task-specific methods. We notice our method shows limited improvements in certain metrics of step-captioning, which is actually  a foreseen situation.  The ultimate goal of HIREST is to train a generic model that can generalize to several related video understanding tasks, so it could  perform worse than certain task-specific methods. 

\subsection{Ablation Study}
We conduct the ablation study to investigate the contribution of each module and the full model. Here, we choose the step-captioning task as the major ablation task, as it requires the model have a comprehensive capability for the cognition, alignment and generation for multi-modal content.
The baseline is \textit{Joint}, which directly fuses three modality representations by element-wise summation and multiplication.


The ablative results are shown in  Table \ref{ablation_module}. The performances of each module and the full QUAG are better than that of \textit{Joint} baseline, which validate their effectiveness. However, the performance improvement is not significant when each module is individually used. When  only QC$^2$ is used, there is even the performance decreasing on the CIDEr metric. Our conjecture is that each module only learns either synergistic relation or filtration relations. Besides, the hierarchies across modalities are overlooked. When the full QUAG model is used, its  performance significantly surpasses the others, in particular with the increases of 20.1\% and 48.7\% on the CIDEr and SPICE over the \textit{Joint} baseline. The ablation results indicate each designed module not only plays its unique role, but also supplements each other. 

\begin{table}[t]
  \centering
  
  \caption{Ablation study for each module on step-captioning.}
    \begin{tabular}{c|cc|cccc}
    \toprule
    Model & MSP & QC$^2$ & M & R & C & S \\
    \midrule
    \textit{Joint}  &   $\times$    & $\times$    & 4.14  & 11.85 & 21.19 & 3.02 \\
    QUAG  &   $\checkmark$     &  $\times$   & 4.70  & 13.05 & 23.18 & 3.49 \\
    QUAG  &     $\times$   &   $\checkmark$    & 4.50  & 12.65 & 20.43 & 3.87 \\
    QUAG  &    $\checkmark$   &     $\checkmark$    & \textbf{4.76} & \textbf{13.20} & \textbf{25.44} & \textbf{4.49} \\
    \bottomrule
    \end{tabular}%
  \label{ablation_module}%
\end{table}%

\begin{table*}[t]
  \centering
  \caption{Performance comparison on the TVsum dataset}
    \begin{tabular}{c|ccccccccccc}
    \toprule
    Model & VT    & VU    & GA    & MS    & PK    & PR    & FM    & BK    & BT    & DS    & Avg. \\
    \midrule 
    TGG \cite{ye2021temporal}  & 85.0  & 71.4  & 81.9  & 78.6  & 80.2  & 75.5  & 71.6  & 77.3  & 78.6  & 68.1  & 76.8 \\
    UMT \cite{liu2022umt}   & 87.5  & 81.5  & 88.2  & 78.8  & 81.4  & \underline{87.0}  & 76.0  & 86.9  & 84.4  & \underline{79.6} & 83.1 \\
    QD-DETR \cite{moon2023query}  & 87.6  & 91.7 & 90.2  & \textbf{88.3} & 84.1  & \textbf{88.3} & \underline{78.7} & 91.2  & 87.8  & 77.7  & 86.6 \\
    UVCOM \cite{xiao2024bridging} & 87.6  & 91.6  & 91.4  & \underline{86.7}  & \underline{86.9}  & 86.9  & 76.9  & 92.3  & 87.4  & 75.6  & 86.3 \\
    TR-DETR \cite{sun2024tr}   & \underline{90.6}  & \underline{92.4}  & \underline{91.7}  & 81.3  & \underline{86.9}  & 85.5  & \textbf{79.8}  & \textbf{93.4}  & \underline{88.3}  & \textbf{81.0} & \textbf{87.1} \\
    \textbf{QUAG} & \textbf{89.8} & \textbf{93.1} & \textbf{92.9} & 85.6 & \textbf{87.9} & 86.1  & 76.7  & 91.4 & \textbf{89.0}  & 77.7  & \underline{87.0} \\
    \bottomrule
    \end{tabular}%
  \label{tvsum}%
\end{table*}%

\begin{figure*}[t]
\centering
\includegraphics[width=1\textwidth]{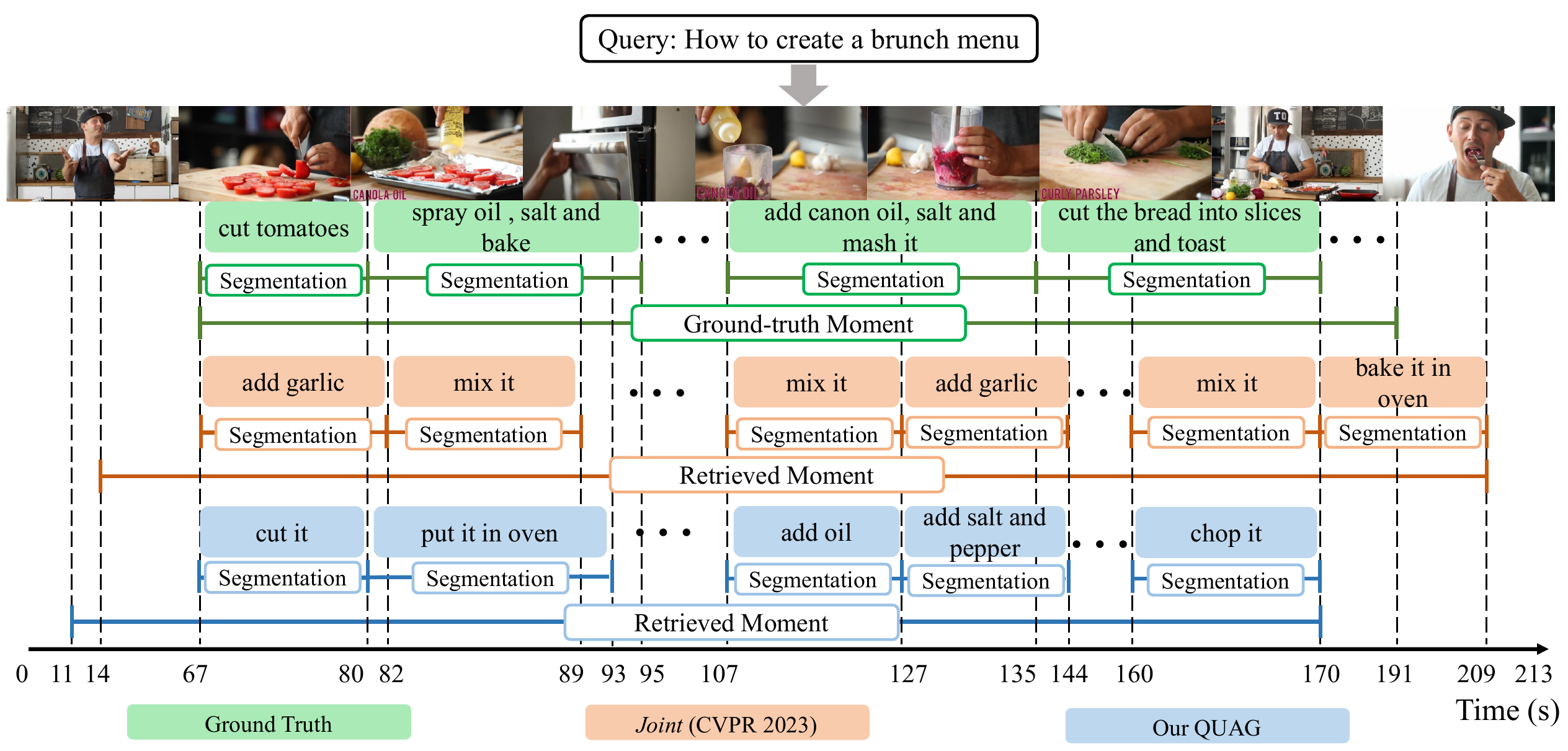} 
  \caption{Given a text query ``How to create a brunch menu'', and the ground-truth annotations, we compare the  predicted and generated outputs of  our QUAG and \textit{Joint}  \cite{zala2023hierarchical}  for moment retrieval, segmentation, and step-captioning. }
\label{vis}
\end{figure*}

\subsection{Performance Comparison on TVsum}
We further validate the generalization ability of QUAG for the query-based video summarization that is related to moment segmentation. The experiment is conducted on the TVsum dataset and the compared methods are TGG \cite{ye2021temporal}, UMT \cite{liu2022umt}, QD-DETR \cite{moon2023query}, UVCOM \cite{xiao2024bridging}, and TR-DETR \cite{sun2024tr}. The inputs for all  compared methods are  visual, audio, and text query representations.

The results are shown in Table \ref{tvsum}. QUAG obtains the best results on the 5 categories, which is better than the others. Besides, the average overall performance of our method is in par with TR-DETR (87.0 vs. 87.1), which shows its good generalization. The superior results mainly benefits from that our method construct a query-centric audio-visual representation. As such, the model can better learn which part of a video is important, and thus extracts the most relevant frames from a video  based on the query.

\subsection{Qualitative Analysis} To obtain an intuitive observation for our method, in Figure \ref{vis}, we provide a visualization case for moment retrieval, segmentation, and step-captioning. 
We observe that the retrieved moment by our QUAG contains the less trivial details than that of the \textit{Joint}. Besides, QUAG better predicts the step boundaries and generates the step captions.  For instance, the yielded step captions by \textit{Joint} model are redundant, which lack the details in the retrieved moment. By contrast, our QUAG generates ``cut it'' that matches the ground-truth caption ``cut tomatoes'' (67-80s); ``add oil'' and ``add salt and pepper'' that show the semantic consistency with the ground-truth caption ``add canon oil, salt and mash it'' (107-135s). Besides, our QUAG can describe fine-grained content of ``chop it'' during the actor cuts the bread into slices. In short, the visualization results show the effectiveness of QUAG. The superiority benefits from the fact of progressively modeling a query-centric audio-visual representation, rather than directly fusing them without distinction. More examples are shown in the supplementary material.


\section{Conclusion}
This paper follows the shallow-to-deep principle to propose QUAG, which learns a query-centric audio-visual representation for moment retrieval, segmentation, and step-captioning. In QUAG, we first devise MSP to gain the audio-visual representation via modeling the global contrastive alignment and local fine-grained interaction between  visual and audio modalities. Then, QC$^2$ is designed to learn a comprehensive cognition for the user-preferred content, by modeling temporal-channel filtration from the deep-level query to shallow-level audio-visual content.  Extensive experiments show QUAG  achieves SOTA performances for the three challenging tasks on the HIREST dataset. Moreover, we test QUAG for  query-based video summarization task. The results  verify its good generalization capability.

\section*{Acknowledgements}
This work was supported by National Natural Science Foundation of China: 62322211, 62336008, U21B2038, ``Pionee'' and ``Leading Goose'' R\&D Program of Zhejiang Province (2024C01023), Fundamental Research Funds for the Central Universities (E2ET1104), Key Laboratory of Intelligent Processing Technology for Digital Music (Zhejiang Conservatory of Music), Ministry of Culture and Tourism (2023DMKLB004).

\bibliography{aaai25}

\section{Supplementary Material}
\subsection{Experiment}
In this supplementary material, we provide the implementation details on the TVsum dataset. In addition, we provide more experimental analyses and discussions about the choice of trade-off parameter. Here, we choose the step-captioning task as the major ablation task, because it requires the model have a comprehensive capability for the cognition, alignment and generation for multi-modal content. Besides, we provide more visualization results and the discussion of failure case.

\subsubsection{Implementation Details}
The proposed QUAG is trained in an end-to-end manner on the TVsum dataset. On TVSum, the clip-level visual features are extracted by using the I3D model \cite{kay2017kinetics} that is pre-trained on Kinetics 400 \cite{carreira2017quo}. The audio features  are extracted by the PANN model \cite{kong2020panns} that is  pre-trained on AudioSet \cite{gemmeke2017audio}. The hidden size is set to 256; the attention layer and attention head are set to 2 and 8, respectively. On TVsum, the batch size and learning rate are set as 4 and 1e-3. The AdamW optimizer \cite{loshchilov2018decoupled} is used to minimize the training loss.  The model training is implemented with PyTorch \cite{paszke2019pytorch} on an RTX 3090 GPU.

\subsubsection{Study on the Trade-off Parameter $\lambda$}
We discuss the effect of trade-off parameter $\lambda$ in Eq. (15) of main paper. As mentioned in the main paper, $\lambda$ is a trade-off parameter to balance the importance between  one of the three task losses and the global contrastive alignment loss. In Table \ref{lambda}, we find that compared to QUAG without the contrastive loss (\emph{i.e.,} setting $\lambda$ to zero), QUAG with the regularization of contrastive loss (under different values) can obtain better results. The results indicate that it is helpful to add the global contrastive alignment loss before modeling the local fine-grained interaction between visual and audio representations. The reason is that this loss can make the visual and audio modalities reside in the same embedding space. 
In addition, it is noted that the results are close when setting the value of $\lambda$ from 0.0001 to 0.0006, and the overall performance of model is better under the value of 0.0003. This shows that the proposed QUAG is robust and not much sensitive to this hyper-parameter. Empirically, we set the value of  $\lambda$ to 0.0003 on the HIREST dataset. In this way, we choose the  value of  $\lambda$ as 0.3 on the TVsum dataset.

\begin{table}[t]
  \centering  
  \small
  \caption{Study on the trade-off parameter $\lambda$ on the step-captioning task, where M, R, C, S are short for METEOR, ROUGE-L, CIDEr, and SPICE.}
    \begin{tabular}{c|c|cccc}
    \toprule
    Model & $\lambda$ & M & R & C & S \\
    \midrule
    QUAG  &   0    & 4.61	& 12.96	& 23.40	& 3.83
 \\
    QUAG  &   0.0001  & 4.71	& 13.17	& 24.08	& \textbf{4.49} \\
    QUAG  &      0.0002    & 4.69	& 13.32	& 25.29	& \textbf{4.49}
 \\  
    QUAG  &     0.0003      & \textbf{4.76} & 13.20 & \textbf{25.44} & \textbf{4.49} \\
    QUAG  &     0.0004     & 4.69	& 13.03	& 24.05	& 4.44
 \\
    QUAG  &     0.0005      & 4.56	& 12.94	& 23.67	& 3.91
 \\
    QUAG  &     0.0006     & 4.60	& \textbf{13.33}	& 25.24	& 4.24
 \\
    \bottomrule
    \end{tabular}%
  \label{lambda}%
\end{table}%

\subsubsection{Ablation Study for Query-Centric Cognition (QC$^2$)}
QC$^2$ aims to use the deep-level query representation to perform a temporal-channel attention on the shallow-level audio-visual representation, thus highlighting user-requested details. The core components in the method are $\mathbf{A}_{te}$ and $\mathbf{A}_{ch}$, which are temporal and channel attention matrices, respectively. The ablation study on $\mathbf{A}_{te}$ and $\mathbf{A}_{ch}$ is shown in Table 4 of the main paper, verifying the effectiveness of using both $\mathbf{A}_{te}$ and $\mathbf{A}_{ch}$. We also investigate the impact of $\mathbf{A}_{te}$ and $\mathbf{A}_{ch}$ by using only one of them. We report the results on step captioning here, where Base replaces $\mathbf{A}_{te}$ and $\mathbf{A}_{ch}$ with the mean-pooling operation. Results show that using either component alone improves Base's performance, and the highest gain is obtained when using both temporal and channel attentions.

\begin{table}[h]
\centering
\caption{Ablation study results on step captioning. M: METEOR, R: ROUGE-L, S: SPICE.}
\begin{tabular}{lccc}
\hline
Ablation & M & R & S \\
\hline
Base & 3.14 & 11.19 & 2.88 \\
QC$^2$ w/o $\mathbf{A}_{ch}$ & 4.28 & 12.09 & 3.56 \\
QC$^2$ w/o $\mathbf{A}_{te}$ & 4.42 & 12.40 & 3.72 \\
QC$^2$  & \textbf{4.50} & \textbf{12.65} & \textbf{3.87} \\
\hline
\end{tabular}
\end{table}

\subsubsection{Qualitative Analysis}
\begin{figure*}[t]
\centering
\includegraphics[width=1\textwidth]{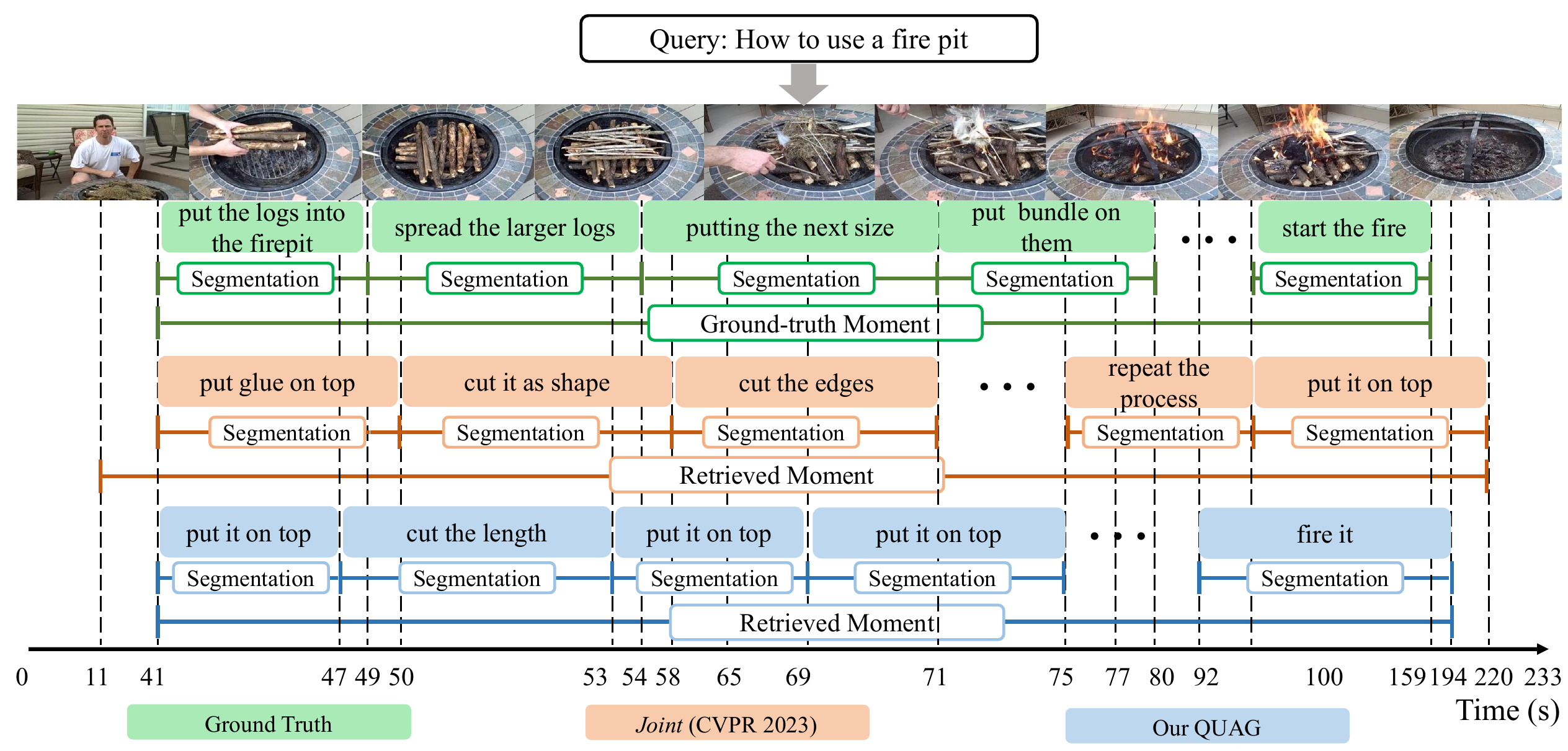} 
  \caption{Given a text query ``How to use a fire pit'', and the ground-truth annotations, we compare the  predicted and generated outputs of  our QUAG and \textit{Joint}  \cite{zala2023hierarchical}  for moment retrieval, segmentation, and step-captioning. }
\label{fig_supp1}
\end{figure*}

\begin{figure*}[t]
\centering
\includegraphics[width=1\textwidth]{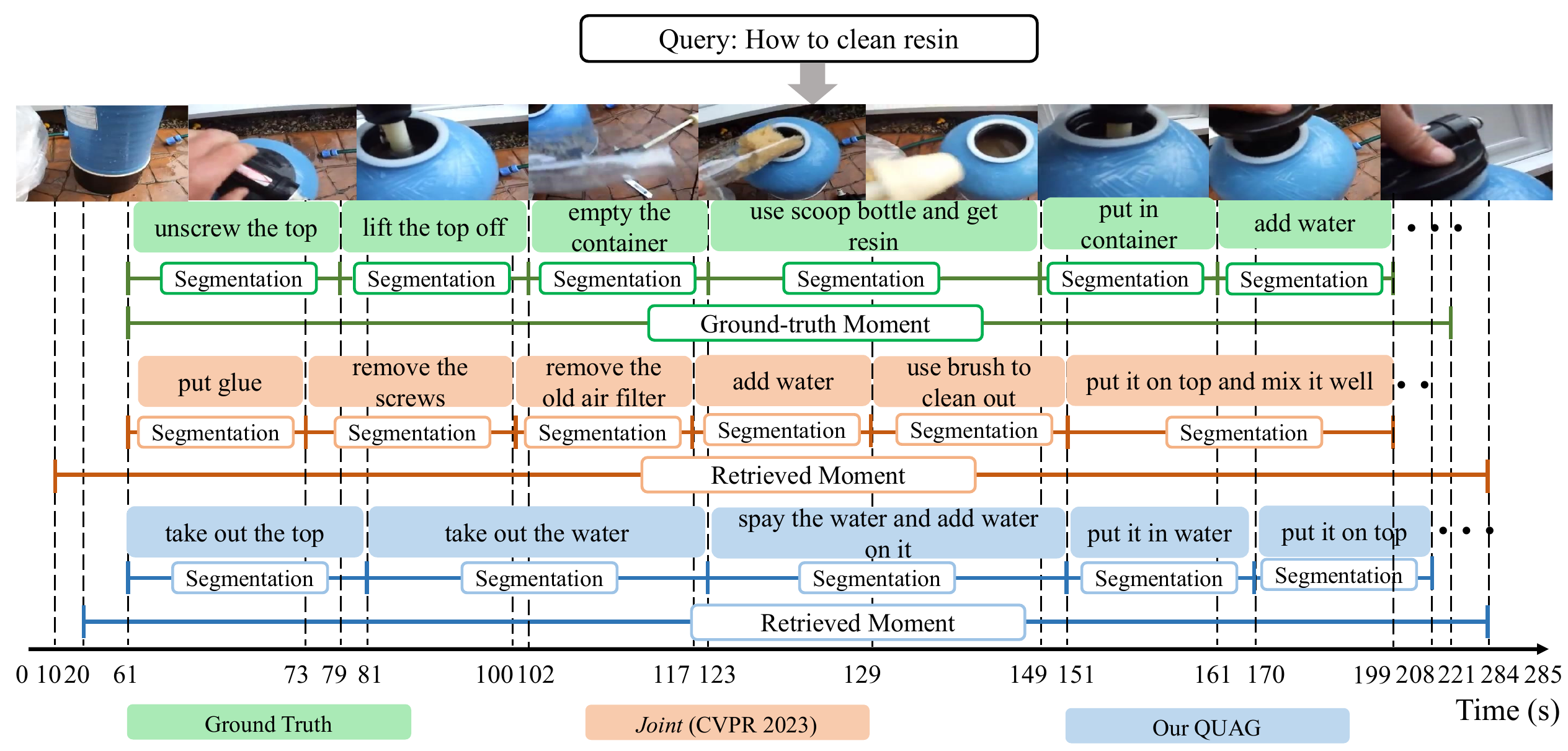} 
  \caption{Given a text query ``How to  clean resin'', and the ground-truth annotations, we compare the  predicted and generated outputs of  our QUAG and \textit{Joint}  \cite{zala2023hierarchical}  for moment retrieval, segmentation, and step-captioning. }
\label{fig_supp2}
\end{figure*}

\begin{figure*}[t]
\centering
\includegraphics[width=1\textwidth]{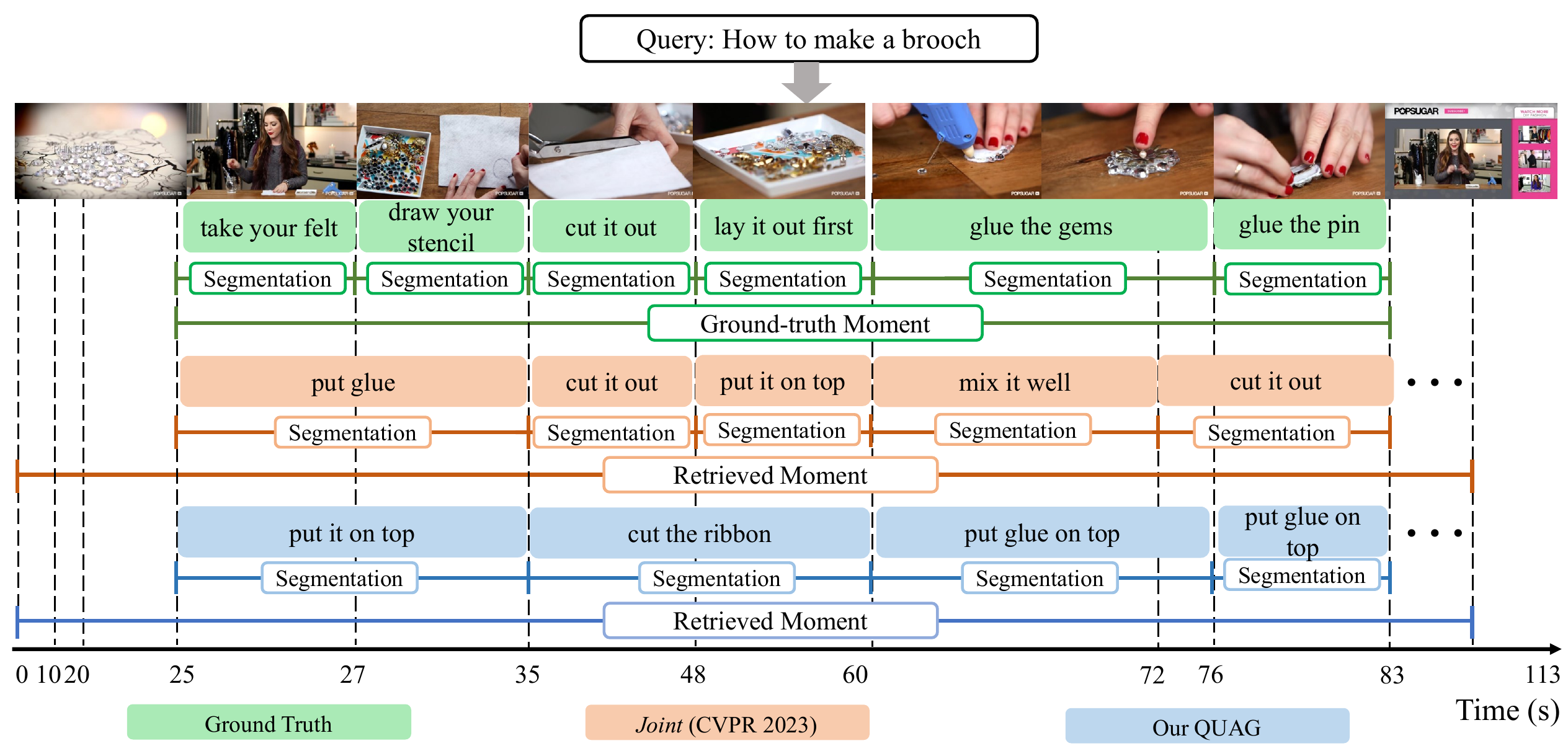} 
  \caption{Given a text query ``How to make a brooch'', and the ground-truth annotations, we compare the  predicted and generated outputs of  our QUAG and \textit{Joint}  \cite{zala2023hierarchical}  for moment retrieval, segmentation, and step-captioning. }
\label{fig_supp3}
\end{figure*}

In this supplementary material, we provide more visualize cases of joint moment retrieval, moment segmentation and step-captioning, which are shown in Figure \ref{fig_supp1}-\ref{fig_supp3}. These cases include  text queries,  ground-truth annotations, and the output results generated by our QUAG and the current state-of-the-art method \textit{Joint} \cite{zala2023hierarchical}. From these cases, we can find that the retrieved moments by our QUAG include the less trivial details than those obtained from the \textit{Joint}. In addition, our QUAG can better predict the step boundaries and generate corresponding step captions.  For instance, in Figure \ref{fig_supp3}, the generated step captions by \textit{Joint} model are redundant, which lack the details in the retrieved moment. By contrast, our QUAG generates ``put glue on top'' that matches the ground-truth captions ``glue the gems'' and ``glue the pin'' (60-83s).  The superiority benefits from the fact that our method can learn a comprehensive cognition for user-preferred video content and thus progressively model  a query-centric audio-visual representation for the three challenging tasks. 

\begin{figure*}[t]
\centering
\includegraphics[width=1\textwidth]{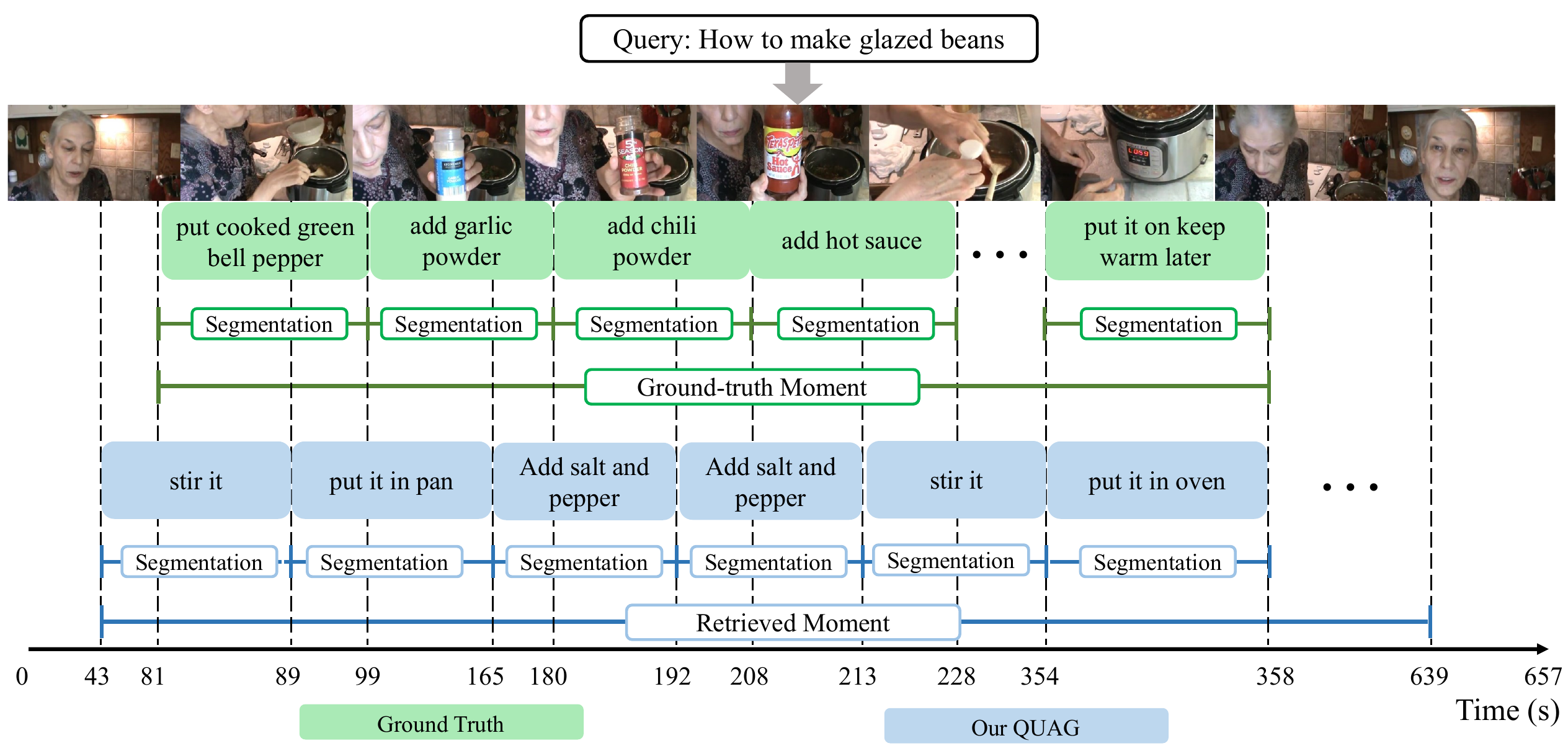} 
  \caption{Given a text query ``How to make glazed beans'', and the ground-truth annotations, we show a failure case of our QUAG for moment retrieval, segmentation, and step-captioning. }
\label{fig_supp_failure}
\end{figure*}

In Figure \ref{fig_supp_failure}, we show a failure case that derives from the proposed QUAG. We find that QUAG localizes the moment with more irrelevant details compared to the ground-truth moment. Our conjecture is that the instructor always appears in the video. In this case, our model would have a misunderstanding that the instructional content has not finished yet.   Besides, although our QUAG can describe some key objects, such as ``oven'', it generates the step captions that fail to describe those fine-grained details, such as ``chili powder'' and ``hot sauce''. For this failure about fine-grained object recognition, a possible solution is to introduce finer-level visual features, such as the segmentation features \cite{kirillov2023segment} to help identify those fine-grained objects.

\end{document}